\documentclass[12pt]{article}       

\usepackage{graphicx}
\usepackage{color}

\usepackage{array}
\usepackage{float}
\usepackage{amsmath}
\usepackage{amssymb}
\usepackage{graphicx}

\providecommand{\tabularnewline}{\\}
\floatstyle{ruled}
\newfloat{algorithm}{tbp}{loa}
\providecommand{\algorithmname}{Algorithm}
\floatname{algorithm}{\protect\algorithmname}

\usepackage[round]{natbib}
\usepackage{hyperref}
\hypersetup{
    colorlinks = true,
    allcolors = {blue}
}
\usepackage{lipsum}
\begin{document}

\title{Fast and Effective Algorithms for Symmetric Nonnegative Matrix Factorization}

\author{Reza Borhani\footnote{R. Borhani and J. Watt contributed equally to this work.}\and
        Jeremy Watt\and
        Aggelos Katsaggelos
}
\date{}
\maketitle

\begin{abstract}
Symmetric Nonnegative Matrix Factorization (SNMF) models arise naturally
as simple reformulations of many standard clustering algorithms including
the popular spectral clustering method. Recent work has demonstrated
that an elementary instance of SNMF provides superior clustering quality
compared to many classic clustering algorithms on a variety of synthetic
and real world data sets. In this work, we present novel reformulations
of this instance of SNMF based on the notion of variable splitting
and produce two fast and effective algorithms for its optimization
using i) the provably convergent Accelerated Proximal Gradient (APG)
procedure and ii) a heuristic version of the Alternating Direction
Method of Multipliers (ADMM) framework.  Our two algorithms present an interesting tradeoff between computational speed and mathematical convergence guarantee: while the former method is provably convergent it is considerably slower than the latter approach, for which we also provide significant but less stringent mathematical proof regarding its convergence.  Through extensive experiments
we show not only that the efficacy of these approaches is equal to
that of the state of the art SNMF algorithm, but also that the latter
of our algorithms is extremely fast being one to two orders of magnitude
faster in terms of total computation time than the state of the art
approach, outperforming even spectral clustering in terms of computation
time on large data sets.
\end{abstract}

\section{Introduction }

In graph-based clustering approaches, data points are treated as individual
nodes in a graph whose edges are weighted using some similarity function.
These weights are stored in a symmetric $n\times n$ adjacency matrix
$\mathbf{A}$ whose $\left(i,j\right)^{th}$ entry\textbf{ $\mathbf{A}_{ij}$}
denotes the similarity between the $i^{th}$ and $j^{th}$ data points
(or nodes)%
\footnote{One common example of a similarity function is the Gaussian similarity
which gives the $\left(i,j\right)^{th}$ entry of $\mathbf{A}$ as
$\mathbf{A}_{ij}=\textrm{exp}\left(-\Vert\mathbf{d}_{i}-\mathbf{d}_{j}\Vert_{2}^{2}/2\sigma^{2}\right)$
where $\mathbf{d}_{i}$ and $\mathbf{d}_{j}$ are the $i^{th}$ and
$j^{th}$ data points, respectively, and $\sigma>0$ is a tuning parameter
that is typically set in proportion to the distribution of the given
data.%
}. A common approach for separating the graph into $K$ clusters is
via an approximate factorization $\mathbf{A}\approx\mathbf{L}\mathbf{L}^{T}$,
where $\mathbf{L}$ is an $n\times K$ nonnegative matrix. The $i^{th}$
data point is then assigned to the $k^{th}$ cluster where $k$ is
the index of the largest entry of the $i^{th}$ row of $\mathbf{L}$.
This factorization is obtained through solving the Symmetric Nonnegative
Matrix Factorization (SNMF) problem defined as
\begin{equation}
\begin{aligned}\underset{\mathbf{L}}{\mbox{minimize}\,\,\,} & \,\,\left\Vert \mathbf{A}-\mathbf{L}\mathbf{L}^{T}\right\Vert _{F}^{2}\\
\mbox{subject to} & \,\,\,\mathbf{L}\geq\mathbf{0},
\end{aligned}
\label{eq:original SNNMF formulation}
\end{equation}

where $\Vert\cdot\Vert_{F}$ denotes the Frobenius norm and the nonnegativity
constraint is taken element-wise. SNMF has been shown to have superior
clustering efficacy compared to a number of data-clustering and graph-clustering
algorithms \citep{kuang2012symmetric, he2011symmetric,yang2012clustering,yang2012quadratic,yang2011unified}. Moreover, slight variations
of the SNMF formulation in (\ref{eq:original SNNMF formulation})
have been shown to be equivalent to a variety of clustering algorithms
including K-means, Nonnegative Matrix Factorization, as well as several
normalized spectral clustering approaches \citep{von2007tutorial,ding2005equivalence,ding2008nonnegative}.
Due to its efficacy and its myriad of connections to other powerful
clustering approaches, efficient algorithms for solving the SNMF model
in (\ref{eq:original SNNMF formulation}) are of particular value
to the practicing data-miner. However, while an array of algorithms
have been developed for the parent problem of Nonnegative Matrix Factorization
(NMF) and NMF-related problems \citep{berry2007algorithms,hoyer2004non,lin2007projected,seung2001algorithms,zhang2010alternating,xu2012alternating},
few specialized algorithms have so far been developed for solving
(\ref{eq:original SNNMF formulation}). This is due to the fact that
the SNMF problem, while having fewer variables to solve for than the
standard NMF formulation, is more challenging to solve due to the
forced equality between the two matrix factors.

\noindent In this paper we develop two novel algorithms for SNMF based
on the notion of variable splitting combined with classic approaches
to constrained numerical optimization, namely, the Quadratic Penalty
Method (QPM) \citep{nocedal2006penalty} with solution via the provably
convergent Accelerated Proximal Gradient (APG) method \citep{beck2009fast,parikh2013proximal},
and a heuristic form of the Alternating Direction Method of Multipliers
(ADMM) framework \citep{boyd2011distributed}. Not only do our algorithms
typically outperform the current state of the art method developed
in \citep{kuang2012symmetric} in terms of clustering efficacy, but
the latter algorithm additionally runs on average one to two orders
of magnitude faster in terms of computation time on medium-sized data
sets consisting of several thousand data points, and can even outperform
spectral clustering in terms of run time on larger data sets.

The remainder of this paper is organized as follows: in the next section
we briefly review popular data- and graph-based clustering approaches
as well as the current state of the art algorithm for solving the
standard SNMF model in (\ref{eq:original SNNMF formulation}). In
section \ref{sec:3} we introduce and derive our two proposed algorithms based
on the notion of variable splitting. We then discuss the computation
time complexity of our fast ADMM based algorithm and compare it to
the state of the art Newton-like procedure in section \ref{sec:4}. The fifth
section contains experiments on synthetic and real data which illustrate
the efficacy and extreme efficiency of our proposed approaches. We
then conclude with brief reflections in section \ref{sec:6}. The appendix of this work then contains critical
mathematical details regarding our first approach, as well as strong
mathematical proof regarding the convergence of our second algorithm.

\section{Review of matrix factorization-based clustering approaches}

In this section we review state of the art data and graph-based clustering
approaches. In addition we highlight many important connections that
exist between this wide array of techniques, which illustrates how
the SNMF problem of interest in this work relates to other methods
via the framework of matrix factorization.

\subsection{Spectral clustering \label{sub:Spectral-clustering}}

Spectral clustering is an immensely popular graph-based clustering
approach that groups $n$ given data points via spectral analysis
of the graph Laplacian matrix $\mathbf{U}$ associated to an input
$n\times n$ adjacency matrix $\mathbf{A}$. This Laplacian is given
by
\begin{equation}
\mathbf{U}=\mathbf{D}-\mathbf{A},
\end{equation}
where $\mathbf{D}$ is the diagonal degree matrix with the $i^{th}$
diagonal entry given by $\mathbf{D}_{ii}=\underset{j=1}{\overset{n}{\sum}}\mathbf{A}_{ij}$,
where $\mathbf{A}_{ij}$ denotes the $\left(i,j\right)^{th}$ entry
of $\mathbf{A}$. The Laplacian matrix $\mathbf{U}$ is symmetric
positive semi-definite, and hence can be diagonalized by an orthogonal
basis of eigenvectors, the top $K$ of which stacked column-wise form a closed form solution to the
unconstrained symmetric matrix factorization problem

\begin{equation}
\begin{aligned}\underset{\mathbf{L}}{\mbox{minimize}\,\,\,} & \,\,\left\Vert \mathbf{U}-\mathbf{L}\mathbf{L}^{T}\right\Vert _{F}^{2}\end{aligned}
.\label{eq:spectral clustering matrix factorization}
\end{equation}

In the spectral clustering framework data is partitioned into $K$
clusters via the $K$ eigenvectors of $\mathbf{U}$ (corresponding
to its smallest $K$ eigenvalues), which are stacked column-wise into
a matrix. The final clustering assignments are then made by performing
K-means on the rows of this matrix \citep{von2007tutorial}.

\noindent A popular normalized version of spectral clustering replaces
$\mathbf{U}$ with a normalized version given by
\begin{equation}
\mathbf{U}\longleftarrow\mathbf{D}^{-1/2}\mathbf{U}\mathbf{D}^{-1/2},
\end{equation}
where $\mathbf{D}^{-1/2}$ denotes the diagonal matrix whose entries
are the square root of the corresponding entries of the inverse of
$\mathbf{D}$, and then follows the same strategy for assigning the
data points to their respective clusters. This simple adjustment to
spectral clustering works significantly better in practice \citep{ng2002spectral}.
Additionally, as mentioned in the introduction section, this normalized
version has very close connections to kernelized K-means and SNMF
(see e.g., Theorem 5 of \citep{ding2005equivalence}).

\subsection{Nonnegative Matrix Factorization (NMF) \label{sub:Nonnegative-Matrix-Factorization}}

\noindent NMF has been shown to be an effective approach for both
dimension reduction and data-based clustering applications. Formally,
in both applications of NMF, we look to recover the factorized approximation
to an $n\times m$ data matrix $\mathbf{H}$ of the form $\mathbf{X}\mathbf{Y}^{T}$
by solving the standard recovery problem below
\begin{equation}
\begin{aligned}\underset{\mathbf{X},\mathbf{Y}}{\mbox{minimize}\,\,\,} & \,\,\left\Vert \mathbf{H}-\mathbf{X}\mathbf{Y}^{T}\right\Vert _{F}^{2}\\
\mbox{subject to} & \,\,\,\mathbf{X},\mathbf{Y}\geq\mathbf{0},
\end{aligned}
\label{eq standard NNMF}
\end{equation}
where $\mathbf{X}$ is an $n\times K$ matrix, $\mathbf{Y}$ is an
$m\times K$ matrix and both are constrained to be nonnegative.
\noindent When employed in clustering applications the matrix $\mathbf{H}$
typically contains the raw data itself \citep{arora2011clustering,berry2007algorithms} and the hypothesized number of clusters $K$ into which the data lies is set
as the number of columns for both matrices $\mathbf{X}$ and $\mathbf{Y}$.  The $i^{th}$ data point is then assigned to the $k^{th}$ cluster
where $k$ is the index of the largest entry of the $i^{th}$ row
of the recovered $\mathbf{Y}$ matrix.
NMF is directly related to the problem of Dictionary Learning, popular
in the signal processing and machine learning communities (see e.g.,
\citep{aharon2005k} and references therein) where a factorization
is desired with the coefficient matrix $\mathbf{Y}$ is constrained
to be sparse. Furthermore both the NMF and Dictionary Learning problems can be thought
of as variations of the basic K-means paradigm, where each column
of $\mathbf{X}$ corresponds to an individual centroid location and
each row of \textbf{$\mathbf{Y}$} is a point's centroid assignment
\citep{aharon2006k,ding2010convex}.

\subsection{Symmetric Nonnegative Matrix Factorization (SNMF) \label{sub:Symmetric-Nonnegative-Matrix}}

The elementary SNMF problem in (\ref{eq:original SNNMF formulation}),
where the matrix $\mathbf{A}$ is an $n\times n$ adjacency matrix,
has been shown to be very effective in graph-based clustering applications.
Using an array of synthetic and real data sets several works  \citep{kuang2012symmetric, he2011symmetric,yang2012clustering,yang2012quadratic,yang2011unified} have shown that algorithms which can solve the SNMF problem produce superior clustering quality compared with standard algorithms including:
normalized spectral clustering as discussed in section \ref{sub:Spectral-clustering},
\textit{\emph{K-means}}, as well as NMF (see section \ref{sub:Nonnegative-Matrix-Factorization}).

One popular and highly effecient algorithm used by \citep{kuang2012symmetric} for solving the
SNMF problem is a simple projected Newton-like method which the authors refer to
as SymNMF. Denoting by $f\left(\mathbf{L}\right)=\left\Vert \mathbf{A}-\mathbf{L}\mathbf{L}^{T}\right\Vert _{F}^{2}$
the objective function of the SNMF problem and $\mathbf{S}$ is an
approximation to the Hessian $\nabla^{2}f\left(\mathbf{L}\right)$,
SymNMF takes projected descent steps of the form

\begin{equation}
\mathbf{x}^{k}=\left[\mathbf{x}^{k-1}-\alpha_{k}\mathbf{S}\nabla f\left(\mathbf{x}^{k-1}\right)\right]^{+}
\end{equation}

where $\alpha_{k}$ is a steplength tuned at each iteration by a standard
adaptive procedure (see e.g., \citep{luenberger2008linear}) to ensure
descent at each step, and the positive part operator $\left[\cdot\right]^{+}$
sets all negative entries of its input to zero. As is always
the case with Newton approximation schemes, the finer $\mathbf{S}$
approximates the true Hessian of $f$ (i.e., as $\mathbf{S}\approx\nabla^{2}f\left(\mathbf{L}\right)$
becomes more accurate) the more rapid is the convergence of the scheme \citep{wright1999numerical},
but the higher the memory and computation overhead of each step. In
their work \citep{kuang2012symmetric} the authors offer several Hessian
approximations schemes that aim at rapid convergence with minimal
overhead.

\section{Proposed methods} \label{sec:3}

In this section we propose two approaches to solving the SNMF problem
in (\ref{eq:original SNNMF formulation}) based on the notion of variable
splitting, an extremely popular reformulation technique in signal
and image processing (see e.g., \citep{goldstein2009split,afonso2010fast,boyd2011distributed}).
In the first instance we propose to solve a quadratic-penalized relaxation
of the original problem, whereas in the second approach we aim at
solving the original problem itself.

\subsection{Variable splitting and relaxation}

Taking the original model in equation (\ref{eq:original SNNMF formulation})
we split the variable $\mathbf{L}$ by introducing a surrogate variable
$\mathbf{Z}$, giving the equivalent problem

\noindent
\begin{equation}
\begin{aligned}\underset{\mathbf{L},\mathbf{Z}}{\mbox{minimize}\,\,\,} & \,\,\left\Vert \mathbf{A}-\mathbf{L}\mathbf{Z}^{T}\right\Vert _{F}^{2}\\
\mbox{subject to} & \,\,\,\mathbf{L},\,\mathbf{Z}\geq\mathbf{0}\\
 & \,\,\,\mathbf{L}-\mathbf{Z}=\mathbf{0}.
\end{aligned}
\label{eq:SNMF 	variable split}
\end{equation}

Note that we have explicitly constrained $\mathbf{Z}\geq\mathbf{0}$,
even though this constraint seems redundant since $\mathbf{L}$ is already constrained to be nonnegative and $\mathbf{Z}$ is constrained to be equal to $\mathbf{L}$.  However it will not be redundant when
we relax the problem by squaring the equality constraint and bringing
it to the objective as

\noindent
\begin{equation}
\begin{aligned}\underset{\mathbf{L},\mathbf{Z}}{\mbox{minimize}\,\,\,} & \,\,\left\Vert \mathbf{A}-\mathbf{L}\mathbf{Z}^{T}\right\Vert _{F}^{2}+\rho\left\Vert \mathbf{L}-\mathbf{Z}\right\Vert _{F}^{2}\\
\mbox{subject to} & \,\,\,\mathbf{L},\,\,\mathbf{Z}\geq\mathbf{0}.
\end{aligned}
\label{eq:QPM reformulation}
\end{equation}

This approach to approximating a constrained optimization problem,
known as the Quadratic Penalty Method (QPM), is widely used in numerical
optimization.  In particular, as $\rho\longrightarrow\infty$ one can show formally that solving this problem is equivalent to solving the constrained problem in (\ref{eq:SNMF variable split}), and hence the original SNMF problem itself shown in (\ref{eq:original SNNMF formulation}).  Generally speaking, however, it is common to set $\rho$ to only a moderate value in practice as this typically provides a solution to the QPM form of a problem that solves the original problem very well for many applications \citep{nocedal2006penalty}.  In our experiments we have found this to be the case for the SNMF problem as well (see Section \ref{sec:experiments}).  Finally, note how this relaxed
form of the SNMF problem can also be thought of as a regularized form of
the standard NMF problem in (\ref{eq standard NNMF}), and is precisely
this problem when $\rho=0$.

\subsection{Accelerated Proximal Gradient (APG) approach}

We solve (\ref{eq:QPM reformulation}) by alternatingly minimizing
over $\mathbf{L}$ and $\mathbf{Z}$, in each case to
convergence, which in turn produces a provably convergent approach
\citep{berry2007algorithms}. In order to do this we employ (in each
direction) the Accelerated Proximal Gradient (APG) method \citep{beck2009fast,parikh2013proximal}.
In the $\mathbf{L}$ direction the standard proximal gradient step,
a descent step projected onto the nonnegative orthant, takes the form
\begin{equation}
\mathbf{L}^{i+1}=\left[\mathbf{L}^{i}-\alpha\left(\left(\mathbf{L}^{i}\left(\mathbf{Z}\right)^{T}-\mathbf{A}\right)\mathbf{Z}+\rho\left(\mathbf{L}^{i}-\mathbf{Z}\right)\right)\right]^{+},\label{eq:up}
\end{equation}

\noindent where the steplength $\alpha$ can be optimally set (see section \ref{sub:Lipschitz-constant-for-APG})
as the reciprocal of the Lipschitz constant of the objective in (\ref{eq:QPM reformulation})
in $\mathbf{L}$ as

\noindent
\begin{equation}
\alpha=\frac{1}{\left\Vert \left(\mathbf{Z}\right)^{T}\mathbf{Z}+\rho\mathbf{I}\right\Vert _{2}}.
\end{equation}
The update procedure in (\ref{eq:up}) is repeated until convergence.
Collecting all terms in $\mathbf{L}^{i}$, (\ref{eq:up}) can be written
in a more computationally efficient manner as

\noindent
\begin{equation}
\mathbf{L}^{i+1}=\left[\mathbf{L}^{i}\left(\left(1-\alpha\rho\right)\mathbf{I}-\alpha\left(\mathbf{Z}\right)^{T}\mathbf{Z}\right)+\alpha\left(\mathbf{A}+\rho\mathbf{I}\right)\mathbf{Z}\right]^{+},
\end{equation}
since the matrices $\left(1-\alpha\rho\right)\mathbf{I}-\alpha\left(\mathbf{Z}\right)^{T}\mathbf{Z}$
and $\alpha\left(\mathbf{A}+\rho\mathbf{I}\right)\mathbf{Z}$ may
be cached and reused at each iteration. Written in this way, the accelerated
form of the proximal gradient step (which is provably an order faster
in terms of convergence%
\footnote{Standard proximal gradient descent is provably convergent to within
$\frac{1}{k}$ of a minimum in $\mathcal{O}\left(k\right)$ iterations,
while APG is convergent to within $\frac{1}{k^{2}}$ in the same order
of steps \citep{beck2009fast,DONE_RIGHT}. %
}) can be written as
\begin{equation}
\begin{aligned}\mathbf{L}^{i+1}=\left[\mathbf{\Phi}^{i}\left(\left(1-\alpha\rho\right)\mathbf{I}-\alpha\left(\mathbf{Z}\right)^{T}\mathbf{Z}\right)+\alpha\left(\mathbf{A}+\rho\mathbf{I}\right)\mathbf{Z}\right]^{+}\\
\mathbf{\Phi}^{i+1}=\mathbf{L}^{i+1}+\frac{i}{i+3}\left(\mathbf{L}^{i+1}-\mathbf{L}^{i}\right).\,\,\,\,\,\,\,\,\,\,\,\,\,\,\,\,\,\,\,\,\,\,\,\,\,\,\,\,\,\,\,\,\,\,\,\,\,\,\,\,\,\,\,\,\,\,\,\,\,\,\,\,\,\,
\end{aligned}
\end{equation}
Using precisely the same ideas, we may write the accelerated proximal
gradient step in $\mathbf{Z}$ as

\noindent
\begin{equation}
\begin{aligned}\mathbf{Z}^{j+1}=\left[\mathbf{\Psi}^{j}\left(\left(1-\beta\rho\right)\mathbf{I}-\beta\left(\mathbf{L}\right)^{T}\mathbf{L}\right)+\beta\left(\mathbf{A}^{T}+\rho\mathbf{I}\right)\mathbf{L}\right]^{+}\\
\mathbf{\Psi}^{j+1}=\mathbf{Z}^{j+1}+\frac{j}{j+3}\left(\mathbf{Z}^{j+1}-\mathbf{Z}^{j}\right),\,\,\,\,\,\,\,\,\,\,\,\,\,\,\,\,\,\,\,\,\,\,\,\,\,\,\,\,\,\,\,\,\,\,\,\,\,\,\,\,\,\,\,\,\,\,\,\,\,\,\,\,\,\,\,\,
\end{aligned}
\end{equation}
where again the steplength $\beta$ may be optimally set (see section \ref{sub:Lipschitz-constant-for-APG}) as the reciprocal
of the Lipschitz constant of (\ref{eq:QPM reformulation}) in $\mathbf{Z}$ as

\noindent
\begin{equation}
\beta=\frac{1}{\left\Vert \left(\mathbf{L}\right)^{T}\mathbf{L}+\rho\mathbf{I}\right\Vert _{2}}.
\end{equation}

For convenience we reproduce the entire alternating accelerated proximal
gradient approach we employ in Algorithm \ref{alg: accelerated proximal gradient},
which we refer to as $\textrm{SNMF}_{\textrm{APG}}$.

\begin{algorithm}[!th]
\textbf{Input: }Adjacency matrix\textbf{ $\mathbf{A}$}, penalty parameter
\textbf{$\rho>0$}, stopping threshold $\epsilon$,

and initializations for $\mathbf{Z}^{0}$, $\mathbf{\Phi}^{0}$, and
$\mathbf{\Psi}^{0}$\textbf{ }

\textbf{Output:} Final point-assignment matrix $\mathbf{Z}^{k}$

\vspace{0.1cm}

$k\leftarrow1$

\textbf{While }$\frac{\left\Vert \mathbf{L}^{k}-\mathbf{L}^{k-1}\right\Vert _{F}}{\left\Vert \mathbf{L}^{k-1}\right\Vert _{F}}+\frac{\left\Vert \mathbf{Z}^{k}-\mathbf{Z}^{k-1}\right\Vert _{F}}{\left\Vert \mathbf{Z}^{k-1}\right\Vert _{F}}>\epsilon$:

~~~~~~(\textit{update} $\mathbf{L}$)

~~~~~~Compute the Lipschitz constant $\alpha_{k}=\frac{1}{\left\Vert \left(\mathbf{Z}^{k-1}\right)^{T}\mathbf{Z}^{k-1}+\rho\mathbf{I}\right\Vert _{2}}$

~~~~~~Reset the counter for the following while loop $i\leftarrow0$

\vspace{0.1cm}

\textbf{~~}~~~~\textbf{While $\frac{\left\Vert \mathbf{L}^{i}-\mathbf{L}^{i-1}\right\Vert _{F}}{\left\Vert \mathbf{L}^{i-1}\right\Vert _{F}}>\epsilon$:}

~~~~~~~~~~~~$\mathbf{R}^{i+1}=\mathbf{\Phi}^{i}\left(\left(1-\alpha_{k-1}\rho\right)\mathbf{I}-\alpha_{k-1}\left(\mathbf{Z}^{k-1}\right)^{T}\mathbf{Z}^{k-1}\right)+\alpha_{k-1}\left(\mathbf{A}+\rho\mathbf{I}\right)\mathbf{Z}^{k-1}$

~~~~~~~~~~~~$\mathbf{L}^{i+1}=\left[\mathbf{R}^{i+1}\right]^{+}$

~~~~~~~~~~~~$\mathbf{\Phi}^{i+1}=\mathbf{L}^{i+1}+\frac{i}{i+3}\left(\mathbf{L}^{i+1}-\mathbf{L}^{i}\right)$

~~~~~~\textbf{End }

\vspace{0.2cm}

~~~~~~$\mathbf{L}^{k}\leftarrow\mathbf{L}^{i+1}$

\vspace{0.4cm}

~~~~~~(\textit{update} $\mathbf{Z}$)

\textbf{~~}~~~~Compute the Lipschitz constant $\beta_{k}=\frac{1}{\left\Vert \left(\mathbf{L}^{k}\right)^{T}\mathbf{L}^{k}+\rho\mathbf{I}\right\Vert _{2}}$

\textbf{~~}~~~~Reset the counter for the following while loop
$j\leftarrow0$

\vspace{0.1cm}

\textbf{~~}~~~~\textbf{While $\frac{\left\Vert \mathbf{Z}^{j}-\mathbf{Z}^{j-1}\right\Vert _{F}}{\left\Vert \mathbf{Z}^{j-1}\right\Vert _{F}}>\epsilon$:}

\textbf{~~}~~\textbf{~~}~~~~~~$\mathbf{S}^{j+1}=\mathbf{\Psi}^{j}\left(\left(1-\beta_{k}\rho\right)\mathbf{I}-\beta_{k}\left(\mathbf{L}^{k}\right)^{T}\mathbf{L}^{k}\right)+\beta\left(\mathbf{A}^{T}+\rho\mathbf{I}\right)\mathbf{L}^{k}$

\textbf{~~}~~\textbf{~~}~~~~~~$\mathbf{Z}^{j+1}=\left[\mathbf{S}^{j+1}\right]^{+}$

\textbf{~~}~~\textbf{~~}~~~~~~$\mathbf{\Psi}^{j+1}=\mathbf{Z}^{j+1}+\frac{j}{j+3}\left(\mathbf{Z}^{j+1}-\mathbf{Z}^{j}\right)$

\textbf{~~}~~~~\textbf{End }

\textbf{~~}~~~~$\mathbf{Z}^{k}\leftarrow\mathbf{Z}^{j+1}$

\textbf{~~}~~~~$k\leftarrow k+1$

\textbf{End }

\caption{\label{alg: accelerated proximal gradient}$\textrm{SNMF}_{\textrm{APG}}$ }
\end{algorithm}

\subsection{Alternating Direction Method of Multipliers approach}

We reformulate problem (\ref{eq:original SNNMF formulation}) slightly
differently than in the previous instance, splitting the variable
$\mathbf{L}$ twice as
\begin{equation}
\begin{aligned}\underset{\mathbf{L},\mathbf{X},\mathbf{Y}}{\mbox{minimize}\,\,\,} & \,\,\frac{1}{2}\left\Vert \mathbf{A}-\mathbf{X}\mathbf{Y}^{T}\right\Vert _{F}^{2}\\
\mbox{subject to} & \,\,\,\mathbf{L}\geq\mathbf{0}\\
 & \,\,\,\mathbf{L}-\mathbf{X}=\mathbf{0}\\
 & \,\,\,\mathbf{L}-\mathbf{Y}=\mathbf{0}.
\end{aligned}
\label{eq:ADMM reformulated problem}
\end{equation}

\noindent This is again an equivalent reformulation of the original
problem in equation (\ref{eq:original SNNMF formulation}). However
unlike the use of APG method previously taken where we aimed to solve
a relaxed form of the SNMF problem, here we aim to solve this reformulated
version of the exact problem itself via a primal-dual method known
as the Alternating Direction Method of Multipliers (ADMM).

\noindent While developed close to a half a century ago, ADMM and
other Lagrange multiplier methods in general have seen an explosion
of recent interest in the machine learning and signal processing communities
\citep{boyd2011distributed,goldstein2009split}. While classically
ADMM has been provably convergent for only convex problems,
recent work has also proven convergence of the method for particular
families of nonconvex problems (see e.g., \citep{zhang2010alternating,xu2012alternating,hong2014convergence,magnusson2014convergence}).
There has also been extensive successful use of ADMM as a heuristic
method for highly nonconvex problems \citep{xu2012alternating,zhang2010alternating,Watt,boyd2011distributed,barman2011decomposition,derbinsky2013improved,fu2013bethe,YOU_ADMM}.
It is in this spirit that we have applied ADMM to our nonconvex problem
and, like these works, find it to provide excellent results empirically
(see section 5). Furthermore, the specific reformulation we have chosen
in (\ref{eq:ADMM reformulated problem}) where we have used two splitting
variables $\mathbf{X}$ and $\mathbf{Y}$, allows us to prove a significant
result regarding convergence of ADMM applied to this reformulation,
i.e., any fixed point of our algorithm is indeed a KKT point
of the original problem (see the Appendix for a proof).  This type of result has in fact been shown to hold when applying ADMM to other matrix factorization problems as well (see e.g., \citep{xu2012alternating,zhang2010alternating}).

Forming the Augmented Lagrangian associated to (\ref{eq:ADMM reformulated problem})
gives
\begin{equation}
\begin{array}{c}
\mathcal{L}\left(\mathbf{X},\mathbf{Y},\mathbf{L},\boldsymbol{\Lambda},\boldsymbol{\Gamma},\rho\right)=\frac{1}{2}\Vert\mathbf{A}-\mathbf{X}\mathbf{Y}^{T}\Vert_{F}^{2}\\
\,\,\,\,\,\,\,\,\,\,\,\,\,\,\,\,\,\,\,\,\,\,\,\,\,\,\,\,\,\,\,\,\,\,\,\,\,\,\,\,\,\,\,\,\,\,\,\,\,\,\,\,\,\,\,\,\,\,\,\,\,\,\,\,\,\,\,\,\,\,+\,\,\frac{\rho}{2}\Vert\mathbf{L}-\mathbf{X}\Vert_{F}^{2}+\left\langle \boldsymbol{\Lambda},\,\mathbf{L}-\mathbf{X}\right\rangle \\
\,\,\,\,\,\,\,\,\,\,\,\,\,\,\,\,\,\,\,\,\,\,\,\,\,\,\,\,\,\,\,\,\,\,\,\,\,\,\,\,\,\,\,\,\,\,\,\,\,\,\,\,\,\,\,\,\,\,\,\,\,\,\,\,\,\,\,\,\,+\,\,\frac{\rho}{2}\Vert\mathbf{L}-\mathbf{Y}\Vert_{F}^{2}+\left\langle \boldsymbol{\Gamma},\,\mathbf{L}-\mathbf{Y}\right\rangle ,
\end{array}\label{eq:Augmented Lagrangian-1}
\end{equation}

\noindent where $\left\langle \cdot,\cdot\right\rangle $ denotes
the inner-product of its input matrices and $\rho>0$ is a parameter that typically requires only a small amount of tuning in practice (see Section \ref{sec:experiments} for further discussion). We will alternate minimizing
$\mathcal{L}$ over primal variables $\mathbf{X}$, $\mathbf{Y}$, and $\mathbf{L}$ with a gradient
ascent step in the dual variables $\boldsymbol{\Lambda}$ and $\boldsymbol{\Gamma}$.
Over $\mathbf{X}$ this reduces to the simple constrained minimization
(after combining terms in $\mathbf{X}$ and ignoring all others) of
the form

\noindent
\begin{equation}
\begin{aligned}\underset{\mathbf{X}}{\mbox{minimize}} & \,\,\frac{1}{2}\left\Vert \mathbf{X}\mathbf{Y}^{T}-\mathbf{A}\right\Vert _{F}^{2}+\frac{\rho}{2}\Vert\mathbf{X}-\left(\mathbf{L}+\frac{1}{\rho}\boldsymbol{\Lambda}\right)\Vert_{F}^{2}\end{aligned}
,
\end{equation}
which is a simple unconstrained quadratic problem. Setting its gradient
to zero gives the optimal solution as
\begin{equation}
\mathbf{X}^{\ast}=\left(\mathbf{A}\mathbf{Y}+\rho\mathbf{L}+\boldsymbol{\Lambda}\right)\left(\mathbf{Y}^{T}\mathbf{Y}+\rho\mathbf{I}\right)^{-1}.
\end{equation}
The invertibility of the rightmost matrix above is assured due to
the addition of the weighted identity $\rho\mathbf{I}$ to $\mathbf{Y}^{T}\mathbf{Y}$.
Similarly, minimizing the Lagrangian in (\ref{eq:Augmented Lagrangian-1})
over $\mathbf{Y}$ reduces to solving another simple quadratic problem
given below

\noindent
\begin{equation}
\begin{aligned}\underset{\mathbf{Y}}{\mbox{minimize}} & \,\,\frac{1}{2}\left\Vert \mathbf{X}\mathbf{Y}^{T}-\mathbf{A}\right\Vert _{F}^{2}+\frac{\rho}{2}\Vert\mathbf{Y}-\left(\mathbf{L}+\frac{1}{\rho}\boldsymbol{\Gamma}\right)\Vert_{F}^{2}\end{aligned}
.
\end{equation}
Again, setting the gradient to zero gives the optimal solution
\begin{equation}
\mathbf{Y}^{\ast}=\left(\mathbf{A}\mathbf{X}+\rho\mathbf{L}+\boldsymbol{\Gamma}\right)\left(\mathbf{X}^{T}\mathbf{X}+\rho\mathbf{I}\right)^{-1}.\label{eq:Y-update}
\end{equation}
Note that in practice rarely do we solve for $\mathbf{Y}^{\ast}$
by actually inverting the matrix $\mathbf{X}^{T}\mathbf{X}+\rho\mathbf{I}$
as in (\ref{eq:Y-update}). Instead, it is more efficient to catch
a Cholesky factorization of this matrix and solve the corresponding
linear system using forward-backward substitution.

\noindent Finally, over $\mathbf{L}$ we have the quadratic minimization
problem with a nonnegativity constraint

\noindent
\begin{equation}
\begin{aligned}\underset{\mathbf{L}}{\mbox{minimize}\,\,\,} & \,\,\left\Vert \mathbf{L}-\left(\mathbf{X}-\frac{1}{\rho}\boldsymbol{\Lambda}\right)\right\Vert _{F}^{2}+\left\Vert \mathbf{L}-\left(\mathbf{Y}-\frac{1}{\rho}\boldsymbol{\Gamma}\right)\right\Vert _{F}^{2}\\
\mbox{subject to} & \,\,\mathbf{L}\geq\mathbf{0}.
\end{aligned}
\end{equation}
Completing the square in $\mathbf{L}$ above, the problem becomes
a projection onto the positive orthant defined by $\mathbf{L}\geq\mathbf{0}$,
whose solution is simply given by

\noindent
\begin{equation}
\mathbf{L}^{\ast}=\frac{1}{2}\left[\mathbf{X}-\frac{1}{\rho}\boldsymbol{\Lambda}+\mathbf{Y}-\frac{1}{\rho}\boldsymbol{\Gamma}\right]^{+}.
\end{equation}
Together with the dual ascent steps we have the full ADMM algorithm
as summarized in Algorithm \ref{alg:ADMM-algorithm-1}, which from
now on is referred to as the $\textrm{SNMF}_{\textrm{ADMM}}$ algorithm.

\begin{algorithm}[!th]
\textbf{Input: }Adjacency matrix\textbf{ $\mathbf{A}$}, penalty parameter
\textbf{$\rho>0$}, stopping threshold $\epsilon$,

and initializations for $\mathbf{Y}^{0}$, $\mathbf{L}^{0}$, $\boldsymbol{\Gamma}^{0}$,
and $\boldsymbol{\Lambda}^{0}$

\textbf{Output:} Final point-assignment matrix $\mathbf{Z}^{k}$

\vspace{0.1cm}

$k\leftarrow1$

\textbf{While $\frac{\left\Vert \mathbf{X}^{k}-\mathbf{X}^{k-1}\right\Vert _{F}}{\left\Vert \mathbf{X}^{k-1}\right\Vert _{F}}+\frac{\left\Vert \mathbf{Y}^{k}-\mathbf{Y}^{k-1}\right\Vert _{F}}{\left\Vert \mathbf{Y}^{k-1}\right\Vert _{F}}+\frac{\left\Vert \mathbf{L}^{k}-\mathbf{L}^{k-1}\right\Vert _{F}}{\left\Vert \mathbf{L}^{k-1}\right\Vert _{F}}>\epsilon$:}

~~~~~~(\textit{update} \textit{primal variable} $\mathbf{X}$)

~~~~~~Find Cholesky factorization of $\left(\mathbf{Y}^{k-1}\right)^{T}\mathbf{Y}^{k-1}+\rho\mathbf{I}\rightarrow\mathbf{C}\mathbf{C}^{T}$

~~~~~~Solve $\mathbf{CJ}=$$\left(\mathbf{A}\mathbf{Y}^{k-1}+\rho\mathbf{L}^{k-1}+\boldsymbol{\Lambda}^{k-1}\right)^{T}$
for $\mathbf{J}$ via forward substitution

~~~~~~Solve $\mathbf{C}^{T}\left(\mathbf{X}^{k}\right)^{T}=$$\mathbf{J}$
for $\mathbf{X}^{k}$ via backward substitution

\vspace{0.4cm}

~~~~~~(\textit{update} \textit{primal variable} $\mathbf{Y}$)

~~~~~~Find Cholesky factorization of $\left(\mathbf{X}^{k}\right)^{T}\mathbf{X}^{k}+\rho\mathbf{I}\rightarrow\mathbf{D}\mathbf{D}^{T}$

~~~~~~Solve $\mathbf{DH}=$$\left(\mathbf{A}\mathbf{X}^{k}+\rho\mathbf{L}^{k-1}+\boldsymbol{\Gamma}^{k-1}\right)^{T}$
for $\mathbf{H}$ via forward substitution

~~~~~~Solve $\mathbf{D}^{T}\left(\mathbf{Y}^{k}\right)^{T}=$$\mathbf{H}$
for $\mathbf{Y}^{k}$ via backward substitution

\vspace{0.4cm}

~~~~~~(\textit{update} \textit{primal variable} $\mathbf{L}$)

~~~~~~$\mathbf{L}^{k}=\frac{1}{2}\left[\mathbf{X}^{k}+\mathbf{Y}^{k}-\frac{1}{\rho}\left(\boldsymbol{\Lambda}^{k-1}+\boldsymbol{\Gamma}^{k-1}\right)\right]^{+}$

\vspace{0.4cm}

~~~~~~(\textit{update dual variable} $\boldsymbol{\Lambda}$)

~~~~~~$\boldsymbol{\Lambda}^{k}=\boldsymbol{\Lambda}^{k-1}+\rho\left(\mathbf{L}^{k}-\mathbf{X}^{k}\right)$

\vspace{0.4cm}

~~~~~~(\textit{update dual variable} $\boldsymbol{\Gamma}$)

~~~~~~$\boldsymbol{\Gamma}^{k}=\boldsymbol{\Gamma}^{k-1}+\rho\left(\mathbf{L}^{k}-\mathbf{Y}^{k}\right)$

~~~~~~$k\leftarrow k+1$

\textbf{End }

\caption{\label{alg:ADMM-algorithm-1}$\textrm{SNMF}_{\textrm{ADMM}}$ }
\end{algorithm}

\section{Time complexity analysis} \label{sec:4}

In this section we compute the per iteration complexity of $\textrm{SNMF}_{\textrm{ADMM}}$
- the fastest of our two proposed algorithms - and compare it to the
per iteration cost of the state of the art SNMF approach mentioned
in section \ref{sub:Symmetric-Nonnegative-Matrix}. As can be seen
in Algorithm \ref{alg:ADMM-algorithm-1}, each iteration of $\textrm{SNMF}_{\textrm{ADMM}}$
includes updating primal variables $\mathbf{X}$, $\mathbf{Y}$, and
$\mathbf{L}$, and dual variables $\boldsymbol{\Lambda}$ and $\boldsymbol{\Gamma}$.
Assuming $\mathbf{Y}\in\mathbb{R}^{n\times K}$, construction of $\mathbf{Y}^{T}\mathbf{Y}+\rho\mathbf{I}$
and corresponding Cholesky factorization, as the first step in updating
$\mathbf{X}$, require approximately $nK^{2}$ and $\frac{1}{3}K^{3}$
operations, respectively. In our analysis we do not account for matrix
(re)assignment operations that can be dealt with via memory pre-allocation.
Additionally, whenever possible we take advantage of the symmetry
of the matrices involved, as is for example the case when computing
$\mathbf{Y}^{T}\mathbf{Y}+\rho\mathbf{I}$. With the sparse graph
structure used in this work (see section \ref{sub:Computational-cost-of FastSNMF} for more
information), the resulting adjacency matrix $\mathbf{A}$ has $q=\left\lfloor log_{2}n\right\rfloor +1$
nonzero entries per row, and therefore computing $\mathbf{A}\mathbf{Y}+\rho\mathbf{L}+\boldsymbol{\Lambda}$
requires approximately $\left(2q+3\right)nK$ operations. Considering
the $2nK^{2}$ operations needed for forward and backward substitutions,
the total per iteration cost of updating $\mathbf{X}$ (or $\mathbf{Y}$)
adds up to $\frac{1}{3}K^{3}+3nK^{2}+2\left(n\, log_{2}n\right)K+5nK$
flops. Updating the primal variable $\mathbf{L}$ can be done using
$6nK$ basic operations. Together with the $3nK$ operations needed
for updating each dual variable, the total per iteration cost of $\textrm{SNMF}_{\textrm{ADMM}}$
is given by

\noindent
\begin{equation}
\tau_{ADMM}=\frac{2}{3}K^{3}+6nK^{2}+4\left(n\, log_{2}n\right)K+22nK.\label{eq:time_ADMM}
\end{equation}

In practice the number of data points $n$ greatly exceeds the number
of clusters $K$, and hence the number of flops in (\ref{eq:time_ADMM})
can be approximated by $\tau_{ADMM}\approx2nK\left(3K+2log_{2}n\right)$.

For comparison the SymNMF algorithm from \citep{kuang2012symmetric},
a projected Newton-like algorithm for solving (\ref{eq:original SNNMF formulation})
(see section \ref{sub:Symmetric-Nonnegative-Matrix}), has comparatively
high per iteration computational cost of $\mathcal{O}\left(n^{3}K^{3}\right)$.
The authors propose a method that takes a limited number of subsampled
Hessian evaluations per iteration. This adjustment lowers the per
iteration cost of their approach to

\begin{equation}
\tau_{SymNMF}\approx\mathcal{O}\left(n^{3}K\right),
\end{equation}

while retaining something of the quadratic convergence of the standard
Newton's method. However, even such an inexpensive Newton's approach
has serious scaling issues in terms of memory and computation time
when dealing with large real world data sets of size $n=10,000$ or
more. Moreover, because $n$ is typically large the per iteration
cost of SymNMF greatly surpasses that of $\textrm{SNMF}_{\textrm{ADMM}}$
derived in (\ref{eq:time_ADMM}).

\section{Experiments} \label{sec:experiments}

In this section we present the results of applying our proposed algorithms
($\textrm{SNMF}_{\textrm{APG}}$ and $\textrm{SNMF}_{\textrm{ADMM}}$)
to several commonly used benchmark data sets including six synthetic
and two real world data sets. In order to evaluate the clustering
efficacy of our algorithm we compare it to the standard normalized
spectral clustering (Spec) algorithm \citep{ng2002spectral}, Nonnegative
Matrix Factorization (NMF) built in MATLAB which uses the popular alternating least squares solution approach \citep{berry2007algorithms,lee2001algorithms}, and to the Symmetric Nonnegative Matrix
Factorization (SymNMF) algorithm \citep{kuang2012symmetric} discussed
in section \ref{sub:Symmetric-Nonnegative-Matrix}. As a stopping
condition for both $\textrm{SNMF}_{\textrm{APG}}$ and $\textrm{SNMF}_{\textrm{ADMM}}$
algorithms a stopping threshold of $\epsilon = 10^{-5}$ was used for all experiments.
This threshold was achieved for all experiments reported here with both algorithms keeping $\rho$ fixed at $\rho=1$ and $\rho=0.1$ for $\textrm{SNMF}_{\textrm{APG}}$ and $\textrm{SNMF}_{\textrm{ADMM}}$, respectively.  These choices were made by running both algorithms on the two real benchmark datasets $5$ times each using $k=20$ clusters in each instance, as detailed in subsection \ref{sub:real-data}, over a set of $20$ equally spaced values for $\rho$ in the range $\left[10^{-2},10\right]$.  We then chose the value of $\rho$ for each algorithm that provided the strongest average performance on the two datasets.  However we note that the choice of $\rho$ was quite robust over the entire range of tested values in these initial experiments for both algorithms.  With each algorithm such a choice for $\rho$ forces the final matrix $\mathbf{L}$ and its respective surrogate variables to be extremely similar when the algorithms converge, and thus the final assignment of datapoints in each experiment (and thus the quality of performance) is calculated as with the original SNMF problem (i.e., the $i^{th}$
data point is assigned to the $k^{th}$ cluster where $k$ is
the index of the largest entry of the $i^{th}$ row of $\mathbf{L}$).  All of the experiments
in this section were run in MATLAB R2012b on a machine with a 3.40
GHz Intel Core i7 processor and 16 GB of RAM.

\subsection{\label{sub:Computational-cost-of FastSNMF} Graph structures }

For large-scale and even medium-sized data sets, it is desirable to
work with sparse adjacency matrices as they significantly lower the
computational time and require less space to store. Therefore, we
follow the suggestion in \citep{von2007tutorial,kuang2012symmetric} and use
sparse graphs. The first step in graph-based clustering of a given
data set $\mathcal{S}=\left\{ \mathbf{d}_{i}\,\vert\,1\leq i\leq n\right\} $,
is to address how to construct the adjacency matrix $\mathbf{A}$.
Following the work in \citep{kuang2012symmetric}, we construct an
adjusted \emph{q-nearest neighbors} graph where the $i^{th}$ data
point (node) is only connected to its \emph{q} nearest neighbors denoted
by $\mathcal{N}_{q}(i)$. The matrix $\mathbf{W}$ then contains the
weights assigned to edges as defined below

\noindent
\begin{equation}
\mathbf{W}_{ij}=\left\{ \begin{array}{cc}
\mbox{exp}\left(-\frac{||\mathbf{d}_{i}-\mathbf{d}_{j}||_{2}^{2}}{\sigma_{i}^{(p)}\sigma_{j}^{(p)}}\right) & \begin{array}{c}
j\in\mathcal{N}_{q}(i)\,\,\textrm{or}\,\, i\in\mathcal{N}_{q}(j)\end{array}\\
0 & \textrm{otherwise}
\end{array}\right.\label{eq:weight}
\end{equation}
Here, the local scale parameter $\sigma_{i}^{(p)}$ is the distance
between $\mathbf{d}_{i}$ and its $p^{th}$ nearest neighbor, where
throughout the experiments $p$ is kept fixed to $7$. As suggested
in \citep{von2007tutorial}, the parameter \emph{q} is chosen as $q=\left\lfloor log_{2}n\right\rfloor +1$
where $n$ is the total number of data points in the data set $\mathcal{S}$.
For the sake of comparability of results, we adopt the normalized
cut objective function used in \citep{kuang2012symmetric} to derive
the adjacency matrix $\mathbf{A}$ from the weight matrix $\mathbf{W}$
via $\mathbf{A}=\mathbf{D}^{-1/2}\mathbf{W}\mathbf{D}^{-1/2}$, where
$\mathbf{D}$ is the diagonal degree matrix associated to $\mathbf{W}$.

\subsection{Synthetic data sets}

\begin{figure}[h]
\noindent \begin{centering}
\includegraphics[scale=0.35]{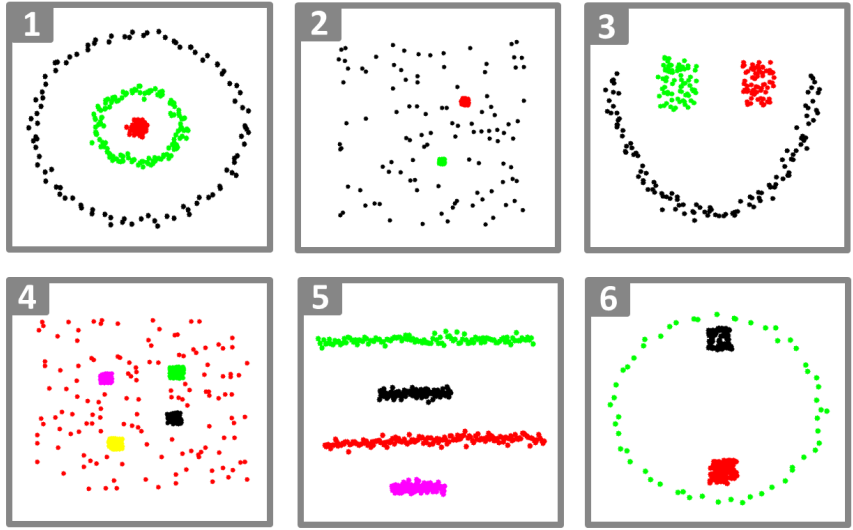}
\par\end{centering}

\caption{\label{fig:Synthetic-datasets.}Synthetic data sets. }
\end{figure}

We first evaluate our algorithms on six synthetic data sets%
\footnote{http://webee.technion.ac.il/\textasciitilde{}lihi/Demos/SelfTuningClustering.html%
} shown in Figure \ref{fig:Synthetic-datasets.}. Each data set is
comprised of 3, 4, or 5 clusters of two-dimensional points and the
total number of data points vary between 238 (data set 6) and 622
(data set 4). All of the algorithms were run 100 times on each data
set with different random initializations for the SNMF factorization
matrix $\mathbf{L}$. At each run the same initialization was used
for SNMF algorithms. Finally, the number of runs at which the algorithms
resulted in perfect clustering of each data set is reported in Table
\ref{tab:Number-of-perfect}, as the measure of clustering performance.
The best result for each data set is highlighted. Based on the results
in Table \ref{tab:Number-of-perfect}, data sets 5 and 6 can be considered
as the most and least challenging data sets, respectively. As can
also be seen our algorithms, particularly $\textrm{SNMF}_{\textrm{ADMM}}$,
perform at least as well, and often outperform SymNMF on these synthetic
data sets. We do not report the results for NMF and K-means algorithms
here since (as expected) both perform very poorly on the types of
data sets shown in Figure \ref{fig:Synthetic-datasets.}.

\begin{table}[h]
\begin{centering}
\begin{tabular}{|>{\centering}m{1.5cm}|>{\centering}m{0.5cm}|>{\centering}m{0.5cm}|>{\centering}m{0.5cm}|>{\centering}m{0.5cm}|}
\multicolumn{1}{>{\centering}m{1.5cm}}{} & \multicolumn{1}{>{\centering}m{0.5cm}}{} & \multicolumn{1}{c}{} & \multicolumn{1}{c}{} & \multicolumn{1}{>{\centering}m{0.5cm}}{}\tabularnewline
\hline
Data

Set  & \multicolumn{1}{c|}{Spec} & \multicolumn{1}{c|}{SymNMF} & \multicolumn{1}{c|}{$\textrm{SNMF}_{\textrm{APG}}$} & \multicolumn{1}{c|}{$\textrm{SNMF}_{\textrm{ADMM}}$}\tabularnewline
\hline
1  & \multicolumn{1}{c|}{66} & \multicolumn{1}{c|}{89} & \multicolumn{1}{c|}{\textbf{90}} & \multicolumn{1}{c|}{89}\tabularnewline
\hline
2  & \multicolumn{1}{c|}{87} & \multicolumn{1}{c|}{\textbf{95}} & \multicolumn{1}{c|}{91} & \multicolumn{1}{c|}{\textbf{95}}\tabularnewline
\hline
3  & \multicolumn{1}{c|}{61} & \multicolumn{1}{c|}{74} & \multicolumn{1}{c|}{75} & \multicolumn{1}{c|}{\textbf{80}}\tabularnewline
\hline
4  & \multicolumn{1}{c|}{62} & \multicolumn{1}{c|}{\textbf{84}} & \multicolumn{1}{c|}{75} & \multicolumn{1}{c|}{77}\tabularnewline
\hline
5  & \multicolumn{1}{c|}{37} & \multicolumn{1}{c|}{71} & \multicolumn{1}{c|}{73} & \multicolumn{1}{c|}{\textbf{80}}\tabularnewline
\hline
6  & \multicolumn{1}{c|}{94} & \multicolumn{1}{c|}{\textbf{100}} & \multicolumn{1}{c|}{\textbf{100}} & \multicolumn{1}{c|}{\textbf{100}}\tabularnewline
\hline
\end{tabular}
\par\end{centering}

\caption{{\small{\label{tab:Number-of-perfect}}} Number of perfect clustering
outputs (out of 100) on the synthetic data sets. The highest score
for each data set is highlighted.}
\end{table}

\subsection{\label{sub:real-data} Real world data sets}

Recently NMF-based algorithms have been extensively used for clustering
tasks, especially for image and document data sets. As representative
of image data sets we use the popular COIL-20%
\footnote{http://www.cs.columbia.edu/CAVE/software/softlib/coil-20.php%
} data set which contains 1,440 128$\times$128 grayscale images consisting of 20 objects,
taken from 72 different viewpoints. Images of the same object then
form a cluster, resulting in 20 equally-sized clusters in total. The
second real data set used here is Reuters-21578%
\footnote{http://www.cad.zju.edu.cn/home/dengcai/Data/TextData.html%
}, a popular text categorization collection consisting of 21,578 documents
manually indexed with categories by the personnel from Reuters Ltd.
In the original Reuters-21578 data set some documents are assigned
more than one label. We remove such documents from the data set so
that each remaining document belongs to only one cluster. Moreover,
we only keep the largest 20 clusters to avoid having clusters with
only a few data points which disrupt the relative balance between
clusters' sizes (these simplifications are commonly made in benchmarking
clustering algorithms using this data set).

\subsection{Evaluation metric}

Upon completion of the cluster assignment process, each cluster is
mapped to one of the gold standard classes. Depending on the size
of the overlap between each cluster and its mapped class, we can then
quantify the efficacy of the clustering algorithm. More formally,
let $c_{i}$ be the cluster label given to the data point $\mathbf{d}_{i}$
by the clustering algorithm, and $g_{i}$ be the provided gold standard
label. The accuracy (AC) score is defined as
\begin{equation}
\textrm{AC}=\frac{\underset{i=1}{\overset{n}{\sum}}\delta[c_{i}-g_{i}]}{n}\times100,
\end{equation}
where $\delta[\cdot]$ is the unit impulse sequence which is 1 at
zero and 0 everywhere else. Here, we use the Kuhn-Munkres algorithm
\citep{matching-book} to find the best mapping. Note that higher
values of AC suggest a better clustering performance, and whenever
the clustering and the gold standard are identical, the accuracy reaches
its maximum, i.e., $\textrm{AC}=100$.

\noindent
\begin{table}[!th]
\begin{centering}
\begin{tabular}{|>{\centering}m{1.5cm}|>{\centering}m{1cm}|>{\centering}m{0.5cm}|>{\centering}m{0.5cm}|>{\centering}m{0.5cm}|>{\centering}m{0.5cm}|}
\multicolumn{1}{>{\centering}m{1.5cm}}{} & \multicolumn{1}{c}{} & \multicolumn{1}{>{\centering}m{0.5cm}}{} & \multicolumn{1}{c}{} & \multicolumn{1}{c}{} & \multicolumn{1}{>{\centering}m{0.5cm}}{}\tabularnewline
\hline
\# of

clusters  & \multicolumn{1}{c|}{NMF} & \multicolumn{1}{c|}{Spec} & \multicolumn{1}{c|}{SymNMF} & \multicolumn{1}{c|}{$\textrm{SNMF}_{\textrm{APG}}$} & \multicolumn{1}{c|}{$\textrm{SNMF}_{\textrm{ADMM}}$}\tabularnewline
\hline
$k=2$  & 89.41  & \multicolumn{1}{c|}{95.24} & \multicolumn{1}{c|}{96.04} & \multicolumn{1}{c|}{96.11} & \multicolumn{1}{c|}{\textbf{96.66}}\tabularnewline
\hline
$k=4$  & 75.54  & \multicolumn{1}{c|}{84.83} & \multicolumn{1}{c|}{\textbf{91.92}} & \multicolumn{1}{c|}{89.55} & \multicolumn{1}{c|}{91.47}\tabularnewline
\hline
$k=6$  & 66.41  & \multicolumn{1}{c|}{79.66} & \multicolumn{1}{c|}{\textbf{83.24}} & \multicolumn{1}{c|}{81.71} & \multicolumn{1}{c|}{81.66}\tabularnewline
\hline
$k=8$  & 64.19  & \multicolumn{1}{c|}{77.88} & \multicolumn{1}{c|}{80.49} & \multicolumn{1}{c|}{79.24} & \multicolumn{1}{c|}{\textbf{81.85}}\tabularnewline
\hline
$\,\,\, k=10$  & 58.78  & \multicolumn{1}{c|}{75.07} & \multicolumn{1}{c|}{80.26} & \multicolumn{1}{c|}{78.47} & \multicolumn{1}{c|}{\textbf{80.28}}\tabularnewline
\hline
$\,\,\, k=12$  & 55.91  & \multicolumn{1}{c|}{69.58} & \multicolumn{1}{c|}{78.85} & \multicolumn{1}{c|}{76.85} & \multicolumn{1}{c|}{\textbf{78.94}}\tabularnewline
\hline
$\,\,\, k=14$  & 53.13  & \multicolumn{1}{c|}{70.16} & \multicolumn{1}{c|}{\textbf{77.65}} & \multicolumn{1}{c|}{76.62} & \multicolumn{1}{c|}{77.38}\tabularnewline
\hline
$\,\,\, k=20$  & \multicolumn{1}{c|}{49.72} & \multicolumn{1}{c|}{60.62} & \multicolumn{1}{c|}{68.39} & \multicolumn{1}{c|}{69.83} & \multicolumn{1}{c|}{\textbf{71.75}}\tabularnewline
\hline
\end{tabular}\\
\begin{tabular}{|>{\centering}m{1.5cm}|>{\centering}m{1cm}|>{\centering}m{0.5cm}|>{\centering}m{0.5cm}|>{\centering}m{0.5cm}|>{\centering}m{0.5cm}|}
\multicolumn{1}{>{\centering}m{1.5cm}}{} & \multicolumn{1}{>{\centering}m{1cm}}{} & \multicolumn{1}{>{\centering}m{0.5cm}}{} & \multicolumn{1}{c}{} & \multicolumn{1}{c}{} & \multicolumn{1}{>{\centering}m{0.5cm}}{}\tabularnewline
\hline
\# of

clusters  & NMF  & \multicolumn{1}{c|}{Spec} & \multicolumn{1}{c|}{SymNMF} & \multicolumn{1}{c|}{$\textrm{SNMF}_{\textrm{APG}}$} & \multicolumn{1}{c|}{$\textrm{SNMF}_{\textrm{ADMM}}$}\tabularnewline
\hline
$k=2$  & 1.07  & \multicolumn{1}{c|}{\textbf{0.02}} & \multicolumn{1}{c|}{0.03} & \multicolumn{1}{c|}{0.42} & \multicolumn{1}{c|}{0.07}\tabularnewline
\hline
$k=4$  & 4.63  & \multicolumn{1}{c|}{\textbf{0.03}} & \multicolumn{1}{c|}{0.16} & \multicolumn{1}{c|}{1.06} & \multicolumn{1}{c|}{0.15}\tabularnewline
\hline
$k=6$  & 7.42  & \multicolumn{1}{c|}{\textbf{0.03}} & \multicolumn{1}{c|}{0.89} & \multicolumn{1}{c|}{2.21} & \multicolumn{1}{c|}{0.27}\tabularnewline
\hline
$k=8$  & 16.55  & \multicolumn{1}{c|}{\textbf{0.05}} & \multicolumn{1}{c|}{2.16} & \multicolumn{1}{c|}{3.35} & \multicolumn{1}{c|}{0.40}\tabularnewline
\hline
$\,\,\, k=10$  & 19.53  & \multicolumn{1}{c|}{\textbf{0.05}} & \multicolumn{1}{c|}{4.60} & \multicolumn{1}{c|}{4.98} & \multicolumn{1}{c|}{0.55}\tabularnewline
\hline
$\,\,\, k=12$  & 24.62  & \multicolumn{1}{c|}{\textbf{0.06}} & \multicolumn{1}{c|}{8.81} & \multicolumn{1}{c|}{7.19} & \multicolumn{1}{c|}{0.71}\tabularnewline
\hline
$\,\,\, k=14$  & 30.92  & \multicolumn{1}{c|}{\textbf{0.07}} & \multicolumn{1}{c|}{16.30} & \multicolumn{1}{c|}{9.59} & \multicolumn{1}{c|}{0.93}\tabularnewline
\hline
$\,\,\, k=20$  & \multicolumn{1}{c|}{82.28} & \multicolumn{1}{c|}{\textbf{0.24}} & \multicolumn{1}{c|}{50.64} & \multicolumn{1}{c|}{18.27} & \multicolumn{1}{c|}{1.53}\tabularnewline
\hline
\end{tabular}
\par\end{centering}

\caption{\label{tab:COIL results}Clustering performance on COIL-20: (top)
clustering efficacy in terms of AC score, and (bottom) computation
time in seconds.}
\end{table}

\begin{table}[!th]
\begin{centering}
\begin{tabular}{|>{\centering}m{1.5cm}|>{\centering}m{1cm}|>{\centering}m{0.5cm}|>{\centering}m{0.5cm}|>{\centering}m{0.5cm}|>{\centering}m{0.5cm}|}
\multicolumn{1}{>{\centering}m{1.5cm}}{} & \multicolumn{1}{c}{} & \multicolumn{1}{>{\centering}m{0.5cm}}{} & \multicolumn{1}{c}{} & \multicolumn{1}{c}{} & \multicolumn{1}{>{\centering}m{0.5cm}}{}\tabularnewline
\hline
\# of

clusters  & \multicolumn{1}{c|}{NMF} & \multicolumn{1}{c|}{Spec} & \multicolumn{1}{c|}{SymNMF} & \multicolumn{1}{c|}{$\textrm{SNMF}_{\textrm{APG}}$} & \multicolumn{1}{c|}{$\textrm{SNMF}_{\textrm{ADMM}}$}\tabularnewline
\hline
$k=2$  & 66.80  & \multicolumn{1}{c|}{91.56} & \multicolumn{1}{c|}{\textbf{92.88}} & \multicolumn{1}{c|}{90.56} & \multicolumn{1}{c|}{91.46}\tabularnewline
\hline
$k=4$  & 47.11  & \multicolumn{1}{c|}{82.99} & \multicolumn{1}{c|}{83.58} & \multicolumn{1}{c|}{82.62} & \multicolumn{1}{c|}{\textbf{83.83}}\tabularnewline
\hline
$k=6$  & 34.95  & \multicolumn{1}{c|}{64.01} & \multicolumn{1}{c|}{73.29} & \multicolumn{1}{c|}{71.50} & \multicolumn{1}{c|}{\textbf{73.58}}\tabularnewline
\hline
$k=8$  & 29.31  & \multicolumn{1}{c|}{58.58} & \multicolumn{1}{c|}{71.26} & \multicolumn{1}{c|}{71.00} & \multicolumn{1}{c|}{\textbf{71.76}}\tabularnewline
\hline
$\,\,\, k=10$  & 29.87  & \multicolumn{1}{c|}{54.85} & \multicolumn{1}{c|}{68.13} & \multicolumn{1}{c|}{67.66} & \multicolumn{1}{c|}{\textbf{68.83}}\tabularnewline
\hline
$\,\,\, k=12$  & 26.87  & \multicolumn{1}{c|}{51.42} & \multicolumn{1}{c|}{68.03} & \multicolumn{1}{c|}{67.14} & \multicolumn{1}{c|}{\textbf{68.06}}\tabularnewline
\hline
$\,\,\, k=14$  & 25.39  & \multicolumn{1}{c|}{45.73} & \multicolumn{1}{c|}{67.05} & \multicolumn{1}{c|}{66.79} & \multicolumn{1}{c|}{\textbf{67.29}}\tabularnewline
\hline
$\,\,\, k=20$  & 23.65  & \multicolumn{1}{c|}{48.92} & \multicolumn{1}{c|}{\textbf{65.80}} & \multicolumn{1}{c|}{64.07} & \multicolumn{1}{c|}{65.76}\tabularnewline
\hline
\end{tabular}\\
\begin{tabular}{|>{\centering}m{1.5cm}|>{\centering}m{1cm}|>{\centering}m{0.5cm}|>{\centering}m{0.5cm}|>{\centering}m{0.5cm}|>{\centering}m{0.5cm}|}
\multicolumn{1}{>{\centering}m{1.5cm}}{} & \multicolumn{1}{>{\centering}m{1cm}}{} & \multicolumn{1}{>{\centering}m{0.5cm}}{} & \multicolumn{1}{c}{} & \multicolumn{1}{c}{} & \multicolumn{1}{>{\centering}m{0.5cm}}{}\tabularnewline
\hline
\# of

clusters  & NMF  & \multicolumn{1}{c|}{Spec} & \multicolumn{1}{c|}{SymNMF} & \multicolumn{1}{c|}{$\textrm{SNMF}_{\textrm{APG}}$} & \multicolumn{1}{c|}{$\textrm{SNMF}_{\textrm{ADMM}}$}\tabularnewline
\hline
$k=2$  & 1.33  & \multicolumn{1}{c|}{0.15} & \multicolumn{1}{c|}{3.67} & \multicolumn{1}{c|}{7.79} & \multicolumn{1}{c|}{\textbf{0.11}}\tabularnewline
\hline
$k=4$  & 2.53  & \multicolumn{1}{c|}{0.23} & \multicolumn{1}{c|}{12.74} & \multicolumn{1}{c|}{14.15} & \multicolumn{1}{c|}{\textbf{0.22}}\tabularnewline
\hline
$k=6$  & 8.64  & \multicolumn{1}{c|}{0.85} & \multicolumn{1}{c|}{104.60} & \multicolumn{1}{c|}{81.43} & \multicolumn{1}{c|}{\textbf{0.84}}\tabularnewline
\hline
$k=8$  & 11.17  & \multicolumn{1}{c|}{1.39} & \multicolumn{1}{c|}{178.66} & \multicolumn{1}{c|}{120.55} & \multicolumn{1}{c|}{\textbf{1.14}}\tabularnewline
\hline
$\,\,\, k=10$  & 11.53  & \multicolumn{1}{c|}{1.29} & \multicolumn{1}{c|}{197.58} & \multicolumn{1}{c|}{121.56} & \multicolumn{1}{c|}{\textbf{1.22}}\tabularnewline
\hline
$\,\,\, k=12$  & 19.53  & \multicolumn{1}{c|}{2.57} & \multicolumn{1}{c|}{416.22} & \multicolumn{1}{c|}{201.20} & \multicolumn{1}{c|}{\textbf{2.18}}\tabularnewline
\hline
$\,\,\, k=14$  & 23.84  & \multicolumn{1}{c|}{2.92} & \multicolumn{1}{c|}{566.93} & \multicolumn{1}{c|}{258.79} & \multicolumn{1}{c|}{\textbf{2.68}}\tabularnewline
\hline
$\,\,\, k=20$  & 31.74  & \multicolumn{1}{c|}{3.97} & \multicolumn{1}{c|}{800.08} & \multicolumn{1}{c|}{383.61} & \multicolumn{1}{c|}{\textbf{3.81}}\tabularnewline
\hline
\end{tabular}
\par\end{centering}

\caption{\label{tab:Reuters}Clustering performance on Reuters-21578: (top)
clustering efficacy in terms of AC score, and (bottom) computation
time in seconds.}
\end{table}

\subsection{Clustering results}

Tables \ref{tab:COIL results} and \ref{tab:Reuters} show the clustering
results on the COIL-20 and Reuters-21578 data sets, respectively.
In these experiments we randomly select $k$ clusters from the entire
data set and run the three clustering algorithms, 5 times each, on
the selected clusters. In each of the 5 instances we feed the same
random initialization for $\mathbf{L}$ to both of our algorithms
as well as SymNMF to fairly compare these algorithms.  Note that spectral
clustering does not require this initialization, and for NMF we use
MATLAB's built-in random initialization. This procedure is repeated
20 times for each $k=2,\,4,\,6\,,\ldots,\,14$ for both data sets.
The last row in both Tables corresponds to the case where the entire
data set is selected ($k=20$). Since there is only one way to select
all clusters, we only report the scores averaged over 5 random initializations
of $\mathbf{L}$.  The computation time reported for all graph-based methods does not include the time spent constructing the neighborhood graph, which for the cases considered here is negligible compared to the runtime of the algorithms themselves.  Moreover since both NMF and the graph-based algorithms must be run several times to ensure a good solution is found, this cost is ameliorated even further over the number of runs of each graph-based algorithm.

The best results are again highlighted in bold. As these Tables show,
both of our algorithms and particularly $\textrm{SNMF}_{\textrm{ADMM}}$,
are highly competitive with the state of the art method (SymNMF) in
terms of clustering quality over all experiments.  That all graph-based approaches outperform NMF is not surprising given their ability to capture a wider array of cluster configuration, as well as their significant engineering advantage over NMF – i.e., they employ a carefully engineered graph transformation of the input data.  In terms of total computation time, while spectral clustering
demonstrates the lowest computation time on the COIL experiments,
$\textrm{SNMF}_{\textrm{ADMM}}$ is significantly faster than SymNMF,
especially for larger values of $k$. Furthermore on the Reuters-21578
experiments, $\textrm{SNMF}_{\textrm{ADMM}}$ is over two orders of
magnitude faster than SymNMF in a majority of the cases, even outperforming
spectral clustering in terms of computation time.

\section{Conclusion} \label{sec:6}

In this work, we have introduced two novel algorithms for solving
the SNMF problem in (\ref{eq:original SNNMF formulation}), an exceptionally
strong model for graph-based clustering applications. In particular,
the experimental evidence put forth in this work indicates that our
algorithms are not only as effective as the state of the art approach,
but also in the case of $\textrm{SNMF}_{\textrm{ADMM}}$
that it may run on average one to two orders of magnitude faster than the state of the art SNMF approach in
terms of computation time. While we make strong statements regarding the mathematical convergence of $\textrm{SNMF}_{\textrm{ADMM}}$ (see Appendix), proving complete convergence of this algorithm remains an open problem for future work.  Thus our two algorithms present an interesting tradeoff between computational speed and mathematical convergence guarantee: $\textrm{SNMF}_{\textrm{APG}}$ is provably convergent yet considerably slower than $\textrm{SNMF}_{\textrm{ADMM}}$, for which less can be currently said regarding provable convergence.  Thus overall the empirical evidence presented here, combined with per iteration complexity analysis and strong proof of mathematical convergence of $\textrm{SNMF}_{\textrm{ADMM}}$, suggests that our algorithms may extend the practical usability of the SNMF framework to general large-scale clustering problems, a research direction we will pursue in the future.

\section{\textcolor{black}{Appendix}}

\subsection{Lipschitz constant for $\textrm{SNMF}_{\textrm{APG}}$ \label{sub:Lipschitz-constant-for-APG}}

If a convex function has Lipschitz continuous gradient with constant
$T$, then the inverse of this value can be used as a fixed step length
for all iterations of (accelerated) proximal gradient (see e.g., \citep{beck2009fast}).
To calculate the Lipschitz constants of $f\left(\mathbf{L},\mathbf{Z}\right)=\left\Vert \mathbf{A}-\mathbf{L}\mathbf{Z}^{T}\right\Vert _{F}^{2}+\rho\left\Vert \mathbf{L}-\mathbf{Z}\right\Vert _{F}^{2}$
in the $\mathbf{L}$ and $\mathbf{Z}$ directions independently it
suffices to compute the maximum eigenvalue of the Hessian in both
directions (see e.g., \citep{DONE_RIGHT}).

In the $\mathbf{L}$ direction we easily have that Hessian in $\mathbf{L}$,
denoted as $\nabla_{\mathbf{L}}^{2}f$, may be written as

\textcolor{black}{
\begin{equation}
\nabla_{\mathbf{L}}^{2}f=\mathbf{Z}^{T}\mathbf{Z}+\rho\mathbf{I},
\end{equation}
}

where $\mathbf{I}$ is the identity matrix of appropriate size. The
Lipschitz constant in this instance is then given as the maximum eigenvalue
of this matrix, i.e., $T_{\mathbf{L}}=\left\Vert \mathbf{Z}^{T}\mathbf{Z}+\rho\mathbf{I}\right\Vert _{2}$.
Likewise in the $\mathbf{Z}$ direction the Hessian in $\mathbf{Z}$,
denoted by $\nabla_{\mathbf{Z}}^{2}f$, may be written as

\textcolor{black}{
\begin{equation}
\nabla_{\mathbf{Z}}^{2}f=\mathbf{L}^{T}\mathbf{L}+\rho\mathbf{I},
\end{equation}
}

and the corresponding Lipschitz constant is then given as $T_{\mathbf{Z}}=\left\Vert \mathbf{L}^{T}\mathbf{L}+\rho\mathbf{I}\right\Vert _{2}$
.

\subsection{Convergence proof of $\textrm{SNMF}_{\textrm{ADMM}}$ to a KKT point }

\textcolor{black}{The Lagrangian corresponding to our reformulated
SNMF problem in (\ref{eq:ADMM reformulated problem}) may be written
as}

\textcolor{black}{
\begin{equation}
\mathcal{L}\left(\mathbf{X},\mathbf{Y},\mathbf{L},\boldsymbol{\Lambda},\boldsymbol{\Gamma},\boldsymbol{\Omega}\right)=\frac{1}{2}\Vert\mathbf{X}\mathbf{Y}^{T}-\mathbf{A}\Vert_{F}^{2}+\left\langle \boldsymbol{\Lambda},\,\mathbf{L}-\mathbf{X}\right\rangle +\left\langle \boldsymbol{\Gamma},\,\mathbf{L}-\mathbf{Y}\right\rangle +\left\langle \boldsymbol{\Omega},\mathbf{L}\right\rangle
\end{equation}
}

\textcolor{black}{where $\boldsymbol{\Omega}\leq0$. The KKT conditions
associated to our problem are then given by}

\textcolor{black}{
\begin{equation}
\begin{array}{c}
\left(\mathbf{X}\mathbf{Y}^{T}-\mathbf{A}\right)\mathbf{Y}-\boldsymbol{\Lambda}=\mathbf{0}\\
\left(\mathbf{Y}\mathbf{X}^{T}-\mathbf{A}\right)\mathbf{X}-\boldsymbol{\Gamma}=\mathbf{0}\\
\boldsymbol{\Lambda}+\boldsymbol{\Gamma}+\boldsymbol{\Omega}=\mathbf{0}\\
\mathbf{X}-\mathbf{L}=\mathbf{0}\\
\mathbf{Y}-\mathbf{L}=\mathbf{0}\\
\mathbf{L}\geq\mathbf{0}\\
\boldsymbol{\Omega}\leq\mathbf{0}\\
\left\langle \boldsymbol{\Omega},\,\mathbf{L}\right\rangle =0
\end{array}
\end{equation}
}

\textcolor{black}{Here the first three equations are given by $\nabla_{\mathbf{X}}\mathcal{L}$,
$\nabla_{\mathbf{Y}}\mathcal{L}$, and $\nabla_{\mathbf{L}}\mathcal{L}$,
the second three equations enforce primal feasibility, the next dual
feasibility, and the final equation ensures complementary slackness
holds. Rearranging the third equation allows us to simplify final
two lines giving the equivalent KKT system:}

\noindent \begin{flushleft}
\textcolor{black}{
\begin{equation}
\begin{array}{c}
\left(\mathbf{X}\mathbf{Y}^{T}-\mathbf{A}\right)\mathbf{Y}-\boldsymbol{\Lambda}=\mathbf{0}\\
\left(\mathbf{Y}\mathbf{X}^{T}-\mathbf{A}\right)\mathbf{X}-\boldsymbol{\Gamma}=\mathbf{0}\\
\mathbf{X}-\mathbf{L}=\mathbf{0}\\
\mathbf{Y}-\mathbf{L}=\mathbf{0}\\
\mathbf{L}\geq\mathbf{0}\\
\boldsymbol{\Lambda}+\boldsymbol{\Gamma}\geq\mathbf{\mathbf{0}}\\
\left\langle \boldsymbol{\Lambda}+\boldsymbol{\Gamma}\,,\,\mathbf{L}\right\rangle =0
\end{array}
\end{equation}
}
\par\end{flushleft}

\textcolor{black}{Let $\mathbf{Z}\overset{\begin{subarray}{c}
\triangle\end{subarray}}{=}\left(\mathbf{X},\mathbf{Y},\mathbf{L},\boldsymbol{\Lambda},\boldsymbol{\Gamma}\right)$ and denote by $\left\{ \mathbf{Z}^{k}\right\} _{k=1}^{\infty}$ the
sequence generated by }$\textrm{SNMF}_{\textrm{ADMM}}$\textcolor{black}{{}
given in algorithm \ref{alg:ADMM-algorithm-1}. Assuming
\begin{equation}
\underset{k\longrightarrow\infty}{lim}\left(\mathbf{Z}^{k+1}-\mathbf{Z}^{k}\right)=0
\end{equation}
then any limit point of $\left\{ \mathbf{Z}^{k}\right\} _{k=1}^{\infty}$
is a KKT point of our reformulation and consequently any limit point
of $\left\{ \mathbf{L}^{k}\right\} _{k=1}^{\infty}$ is a KKT point
of the original SNMF problem.}

\textcolor{black}{Subtracting off the previous iterate from each line
of our ADMM algorithm gives the following set of equations}

\textcolor{black}{
\begin{equation}
\begin{array}{c}
\mathbf{X}^{k+1}-\mathbf{X}^{k}=\left(\mathbf{A}\mathbf{Y}^{k}+\rho\mathbf{L}^{k}+\boldsymbol{\Lambda}^{k}\right)\left(\left(\mathbf{Y}^{k}\right)^{T}\mathbf{Y}^{k}+\rho\mathbf{I}\right)^{-1}-\mathbf{X}^{k}\\
\mathbf{Y}^{k+1}-\mathbf{Y}^{k}=\left(\mathbf{A}\mathbf{X}^{k+1}+\rho\mathbf{L}^{k}+\boldsymbol{\Gamma}^{k}\right)\left(\left(\mathbf{X}^{k}\right)^{T}\mathbf{X}^{k}+\rho\mathbf{I}\right)^{-1}-\mathbf{Y}^{k}\\
\mathbf{L}^{k+1}-\mathbf{L}^{k}=\frac{1}{2}\left[\mathbf{X}^{k+1}-\frac{1}{\rho}\boldsymbol{\Lambda}^{k}+\mathbf{Y}^{k+1}-\frac{1}{\rho}\boldsymbol{\Gamma}^{k}\right]^{+}-\mathbf{L}^{k}\\
\boldsymbol{\Lambda}^{k+1}-\boldsymbol{\Lambda}^{k}=\rho\left(\mathbf{L}^{k+1}-\mathbf{X}^{k+1}\right)\\
\boldsymbol{\Gamma}^{k+1}=\rho\left(\mathbf{L}^{k+1}-\mathbf{Y}^{k+1}\right)
\end{array}
\end{equation}
}

\textcolor{black}{Since we have assumed $\mathbf{Z}^{k+1}-\mathbf{Z}^{k}\longrightarrow\mathbf{0}$
the left (and consequently right) hand side of each equation above
goes to zero. Isolating the right hand side of each and simplifying
gives }

\textcolor{black}{
\begin{equation}
\begin{array}{c}
\left(\mathbf{X}^{k}\left(\mathbf{Y}^{k}\right)^{T}-\mathbf{A}\right)\mathbf{Y}^{k}-\boldsymbol{\Lambda}^{k}=\mathbf{0}\\
\left(\mathbf{Y}^{k}\left(\mathbf{X}^{k}\right)^{T}-\mathbf{A}\right)\mathbf{X}^{k}-\boldsymbol{\Gamma}^{k}=\mathbf{0}\\
\frac{1}{2}\left[\mathbf{X}^{k+1}-\frac{1}{\rho}\boldsymbol{\Lambda}^{k}+\mathbf{Y}^{k+1}-\frac{1}{\rho}\boldsymbol{\Gamma}^{k}\right]^{+}-\mathbf{L}^{k}=\mathbf{0}\\
\mathbf{L}^{k+1}-\mathbf{X}^{k+1}=\mathbf{0}\\
\mathbf{L}^{k+1}-\mathbf{Y}^{k+1}=\mathbf{0}
\end{array}
\end{equation}
}

\textcolor{black}{From the first and last two equations we can see
that the first four KKT conditions are satisfied at any limit point
of the ADMM algorithm}

\textcolor{black}{
\begin{equation}
\hat{\mathbf{Z}}=\left(\hat{\mathbf{X}},\hat{\mathbf{Y}},\hat{\mathbf{L}},\hat{\boldsymbol{\Lambda}},\hat{\boldsymbol{\Gamma}}\right)
\end{equation}
}

\textcolor{black}{In particular at such a limit point we have, by
construction of the algorithm, that $\hat{\mathbf{L}}\geq\mathbf{0}$.
Since $\hat{\mathbf{L}}=\hat{\mathbf{X}}=\hat{\mathbf{Y}}$ some simple
substitution into the third ADMM equation gives}

\textcolor{black}{
\begin{equation}
\left[\hat{\mathbf{L}}-\frac{1}{2\rho}\left(\hat{\boldsymbol{\Lambda}}+\hat{\boldsymbol{\Gamma}}\right)\right]^{+}=\hat{\mathbf{L}}
\end{equation}
}

\textcolor{black}{Now if $\hat{\mathbf{L}}=\mathbf{0}$ then we have
$\left[-\left(\hat{\boldsymbol{\Lambda}}+\hat{\boldsymbol{\Gamma}}\right)\right]^{+}=0$
in which case $\hat{\boldsymbol{\Lambda}}+\hat{\boldsymbol{\Gamma}}\geq\mathbf{0}$.
Otherwise, if $\hat{\mathbf{L}}>\mathbf{0}$ then it must be the case
that $\hat{\boldsymbol{\Lambda}}+\hat{\boldsymbol{\Gamma}}=\mathbf{0}$.
This shows at any limit point $\hat{Z}$ that the final KKT conditions
($\hat{\boldsymbol{\Lambda}}+\hat{\boldsymbol{\Gamma}}\geq\mathbf{0}$
and complementary) hold as well. Hence we have shown that the sequence
$\left\{ \mathbf{Z}^{k}\right\} _{k=1}^{\infty}$ indeed converges
to a KKT point of the reformulation in (\ref{eq:ADMM reformulated problem}).
The fact that $\left\{ \mathbf{L}^{k}\right\} _{k=1}^{\infty}$ converges
to a KKT point of the original SNMF formulation in (\ref{eq:original SNNMF formulation})
follows immediately from equivalence of our reformulation to this
problem. This shows that - when convergent - the output of our ADMM
algorithm is a KKT point of the original SNMF problem in (\ref{eq:original SNNMF formulation}).}

\bibliographystyle{spbasic}
\bibliography{SNMF}

\begin{thebibliography}{37}
\providecommand{\natexlab}[1]{#1}
\providecommand{\url}[1]{{#1}}
\providecommand{\urlprefix}{URL }
\expandafter\ifx\csname urlstyle\endcsname\relax
  \providecommand{\doi}[1]{DOI~\discretionary{}{}{}#1}\else
  \providecommand{\doi}{DOI~\discretionary{}{}{}\begingroup
  \urlstyle{rm}\Url}\fi
\providecommand{\eprint}[2][]{\url{#2}}

\bibitem[{Afonso et~al(2010)Afonso, Bioucas-Dias, and
  Figueiredo}]{afonso2010fast}
Afonso MV, Bioucas-Dias JM, Figueiredo MA (2010) Fast image recovery using
  variable splitting and constrained optimization. Image Processing, IEEE
  Transactions on 19(9):2345--2356

\bibitem[{Aharon et~al(2005)Aharon, Elad, and Bruckstein}]{aharon2005k}
Aharon M, Elad M, Bruckstein A (2005) K-svd and its non-negative variant for
  dictionary design. In: Proc. of the SPIE Conf. Wavelets, vol 5914

\bibitem[{Aharon et~al(2006)Aharon, Elad, and Bruckstein}]{aharon2006k}
Aharon M, Elad M, Bruckstein A (2006) K-svd: An algorithm for designing
  overcomplete dictionaries for sparse representation. IEEE Trans on Signal
  Processing 54(11):4311

\bibitem[{Arora et~al(2011)Arora, Gupta, Kapila, and
  Fazel}]{arora2011clustering}
Arora R, Gupta M, Kapila A, Fazel M (2011) Clustering by left-stochastic matrix
  factorization. In: Proc. of the 28th Int'l Conf. on Machine Learning, pp
  761--768

\bibitem[{Barman et~al(2011)Barman, Liu, Draper, and
  Recht}]{barman2011decomposition}
Barman S, Liu X, Draper S, Recht B (2011) Decomposition methods for large scale
  lp decoding. In: Communication, Control, and Computing (Allerton), 2011 49th
  Annual Allerton Conference on, IEEE, pp 253--260

\bibitem[{Beck and Teboulle(2009)}]{beck2009fast}
Beck A, Teboulle M (2009) A fast iterative shrinkage-thresholding algorithm for
  linear inverse problems. SIAM Journal on Imaging Sciences 2(1):183--202

\bibitem[{Berry et~al(2007)Berry, Browne, Langville, Pauca, and
  Plemmons}]{berry2007algorithms}
Berry M, Browne M, Langville A, Pauca P, Plemmons R (2007) Algorithms and
  applications for approximate nonnegative matrix factorization. Computational
  Statistics \& Data Analysis 52(1):155--173

\bibitem[{Boyd et~al(2011)Boyd, Parikh, Chu, Peleato, and
  Eckstein}]{boyd2011distributed}
Boyd S, Parikh N, Chu E, Peleato B, Eckstein J (2011) Distributed optimization
  and statistical learning via the alternating direction method of multipliers.
  Foundations and Trends{\textregistered} in Machine Learning 3(1):1--122

\bibitem[{Derbinsky et~al(2013)Derbinsky, Bento, Elser, and
  Yedidia}]{derbinsky2013improved}
Derbinsky N, Bento J, Elser V, Yedidia JS (2013) An improved three-weight
  message-passing algorithm. arXiv preprint arXiv:13051961

\bibitem[{Ding et~al(2005)Ding, He, and Simon}]{ding2005equivalence}
Ding C, He X, Simon H (2005) On the equivalence of nonnegative matrix
  factorization and spectral clustering. In: Proc. of the Fifth SIAM Int'l
  Conf. on Data Mining, vol~5, pp 606--610

\bibitem[{Ding et~al(2008)Ding, Li, and Jordan}]{ding2008nonnegative}
Ding C, Li T, Jordan M (2008) Nonnegative matrix factorization for
  combinatorial optimization: Spectral clustering, graph matching, and clique
  finding. In: 8th IEEE Int'l Conf. on Data Mining, pp 183--192

\bibitem[{Ding et~al(2010)Ding, Li, and Jordan}]{ding2010convex}
Ding C, Li T, Jordan M (2010) Convex and semi-nonnegative matrix
  factorizations. IEEE Trans on Pattern Analysis and Machine Intelligence
  32(1):45--55

\bibitem[{Fu and Banerjee(2013)}]{fu2013bethe}
Fu Q, Banerjee HWA (2013) Bethe-admm for tree decomposition based parallel map
  inference. In: Conference on Uncertainty in Artificial Intelligence (UAI)

\bibitem[{Goldstein and Osher(2009)}]{goldstein2009split}
Goldstein T, Osher S (2009) The split bregman method for l1-regularized
  problems. SIAM Journal on Imaging Sciences 2(2):323--343

\bibitem[{He et~al(2011)He, Xie, Zdunek, Zhou, and Cichocki}]{he2011symmetric}
He Z, Xie S, Zdunek R, Zhou G, Cichocki A (2011) Symmetric nonnegative matrix
  factorization: Algorithms and applications to probabilistic clustering.
  Neural Networks, IEEE Transactions on 22(12):2117--2131

\bibitem[{Hong et~al(2014)Hong, Luo, and Razaviyayn}]{hong2014convergence}
Hong M, Luo ZQ, Razaviyayn M (2014) Convergence analysis of alternating
  direction method of multipliers for a family of nonconvex problems. arXiv
  preprint arXiv:14101390

\bibitem[{Hoyer(2004)}]{hoyer2004non}
Hoyer P (2004) Non-negative matrix factorization with sparseness constraints.
  The Journal of Machine Learning Research 5:1457--1469

\bibitem[{Kuang et~al(2012)Kuang, Park, and Ding}]{kuang2012symmetric}
Kuang D, Park H, Ding C (2012) Symmetric nonnegative matrix factorization for
  graph clustering. In: Proc. of SIAM Data Mining Conference, vol~12, pp
  106--117

\bibitem[{Lee and Seung(2001)}]{lee2001algorithms}
Lee DD, Seung HS (2001) Algorithms for non-negative matrix factorization. In:
  Advances in neural information processing systems, pp 556--562

\bibitem[{Lin(2007)}]{lin2007projected}
Lin C (2007) Projected gradient methods for nonnegative matrix factorization.
  Neural computation 19(10):2756--2779

\bibitem[{Lovasz and Plummer(2009)}]{matching-book}
Lovasz L, Plummer M (2009) Matching Theory. AMS Chelsea Publishing

\bibitem[{Luenberger and Ye(2008)}]{luenberger2008linear}
Luenberger DG, Ye Y (2008) Linear and nonlinear programming, vol 116. Springer

\bibitem[{von Luxburg(2007)}]{von2007tutorial}
von Luxburg U (2007) A tutorial on spectral clustering. Statistics and
  computing 17(4):395--416

\bibitem[{Magn{\'u}sson et~al(2014)Magn{\'u}sson, Weeraddana, Rabbat, and
  Fischione}]{magnusson2014convergence}
Magn{\'u}sson S, Weeraddana PC, Rabbat MG, Fischione C (2014) On the
  convergence of alternating direction lagrangian methods for nonconvex
  structured optimization problems. arXiv preprint arXiv:14098033

\bibitem[{Ng et~al(2002)Ng, Jordan, and Weiss}]{ng2002spectral}
Ng AY, Jordan MI, Weiss Y (2002) On spectral clustering: Analysis and an
  algorithm. Advances in neural information processing systems 2:849--856

\bibitem[{Nocedal and Wright(2006)}]{nocedal2006penalty}
Nocedal J, Wright SJ (2006) Penalty and Augmented Lagrangian Methods. Springer

\bibitem[{Parikh and Boyd(2013)}]{parikh2013proximal}
Parikh N, Boyd SP (2013) Proximal algorithms. Foundations and Trends in
  Optimization pp 1--96

\bibitem[{Seung and Lee(2001)}]{seung2001algorithms}
Seung H, Lee D (2001) Algorithms for non-negative matrix factorization.
  Advances in neural information processing systems 13:556--562

\bibitem[{Watt et~al(2014)Watt, Borhani, and Katsaggelos}]{Watt}
Watt J, Borhani R, Katsaggelos A (2014) A fast, effective, and scalable
  algorithm for symmetric nonnegative matrix factorization. NU Technical Report

\bibitem[{Watt et~al(2016)Watt, Borhani, and Katsaggelos}]{DONE_RIGHT}
Watt J, Borhani R, Katsaggelos A (2016) Machine Learning Refined: Foundations,
  Algorithms, and Applications. Cambridge University Press

\bibitem[{Wright and Nocedal(1999)}]{wright1999numerical}
Wright SJ, Nocedal J (1999) Numerical optimization, vol~2. Springer New York

\bibitem[{Xu et~al(2012)Xu, Yin, Wen, and Zhang}]{xu2012alternating}
Xu Y, Yin W, Wen Z, Zhang Y (2012) An alternating direction algorithm for
  matrix completion with nonnegative factors. Frontiers of Mathematics in China
  7(2):365--384

\bibitem[{Yang and Oja(2011)}]{yang2011unified}
Yang Z, Oja E (2011) Unified development of multiplicative algorithms for
  linear and quadratic nonnegative matrix factorization. Neural Networks, IEEE
  Transactions on 22(12):1878--1891

\bibitem[{Yang and Oja(2012)}]{yang2012quadratic}
Yang Z, Oja E (2012) Quadratic nonnegative matrix factorization. Pattern
  Recognition 45(4):1500--1510

\bibitem[{Yang et~al(2012)Yang, Hao, Dikmen, Chen, and
  Oja}]{yang2012clustering}
Yang Z, Hao T, Dikmen O, Chen X, Oja E (2012) Clustering by nonnegative matrix
  factorization using graph random walk. In: Advances in Neural Information
  Processing Systems, pp 1079--1087

\bibitem[{You and Peng(2014)}]{YOU_ADMM}
You S, Peng Q (2014) A non-convex alternating direction method of multipliers
  heuristic for optimal power flow. IEEE SmartGridComm (to appear)

\bibitem[{Zhang(2010)}]{zhang2010alternating}
Zhang Y (2010) An alternating direction algorithm for nonnegative matrix
  factorization. preprint

\end{thebibliography}
\end{document}